\documentclass[]{spie}  

\usepackage{subfigure}
\usepackage{amsmath,amsfonts,amssymb}
\usepackage{graphicx}
\usepackage{float}
\usepackage[colorlinks=true, allcolors=blue]{hyperref}
\usepackage{longtable}
\usepackage{multicol}
\usepackage{multirow}
\title{Effective 3D Humerus and Scapula Extraction using Low-contrast and High-shape-variability MR Data}

\author[a]{Xiaoxiao He}
\author[a]{Chaowei Tan}
\author[b]{Yuting Qiao}
\author[c]{Virak Tan}
\author[a]{Dimitris Metaxas}
\author[a,c]{Kang Li}
\affil[a]{Department of Computer Science, Rutgers University, Piscataway, USA}
\affil[b]{Department of Industrial and Systems Engineering, Rutgers University, Piscataway, USA}
\affil[c]{Department of Orthopaedics, New Jersey Medical School, Rutgers University, Newark, USA}

\authorinfo{Further author information:  (Send correspondence to Kang Li)\\
Xiaoxiao He: Email: xh172@scarletmail.rutgers.edu\\
Chaowei Tan: Email: ct382@cs.rutgers.edu\\
Yuting Qiao: Email: yq44@scarletmail.rutgers.edu\\
Virak Tan: Email: vtan@handarmsurgery.com\\
Dimitris Metaxas: Email: dnm@cs.rutgers.edu\\
Kang Li: Email: kang.li.research@gmail.com}

\begin{document} 
\maketitle

\begin{abstract}
For the initial shoulder preoperative diagnosis, it is essential to obtain a three-dimensional (3D) bone mask from medical images, e.g., magnetic resonance (MR). However, obtaining high-resolution and dense medical scans is both costly and time-consuming. In addition, the imaging parameters for each 3D scan may vary from time to time and thus increase the variance between images. Therefore, it is practical to consider the bone extraction on low-resolution data which may influence imaging contrast and make the segmentation work difficult. In this paper, we present a joint segmentation for the humerus and scapula bones on a small dataset with low-contrast and high-shape-variability 3D MR images. The proposed network has a deep end-to-end architecture to obtain the initial 3D bone masks. Because the existing scarce and inaccurate human-labeled ground truth, we design a self-reinforced learning strategy to increase performance. By comparing with the non-reinforced segmentation and a classical multi-atlas method with joint label fusion, the proposed approach obtains better results.
\end{abstract}

\keywords{Deep end-to-end network, Self-reinforced learning, Humerus and scapula segmentation}

\section{INTRODUCTION}
\label{sec:intro}  

Multiple analysis and quantification tasks on evaluating shoulder instability and planning shoulder preoperative diagnosis are based on the analysis of the structure and morphology of human humerus and scapula from the volumetric images that have been generated by computed tomography (CT) or magnetic resonance (MR) imaging. In order to prepare an accurate preoperative diagnosis of humerus and scapula for evaluating pathologies, the volumetric data first needs to be segmented, and then the structural and morphological analysis of the two bones can be applied\cite{chuang2008use,acid2012preoperative}. Manually labeling the humerus and scapula data in three-dimensional (3D) space is a tedious, laborious, and time-consuming task. A 3D CT/MR image is captured by combining 2D slices along a specific scanning direction. Therefore, the manual labeling on a 3D image is actually the labeling of multiple 2D images, which costs much time and efforts. Moreover, it is hard for humans to take spatial characteristics, including the relation between slices and the continuity in shape from different axes, into consideration during manually labeling 2D slices. Thus an effective automatic 3D segmentation algorithm to extract the humerus and scapula is necessary and can offer reliability and repeatability in various medical applications.

\begin{figure}[!t]
\begin{center}
    \begin{tabular}{c} 
    \includegraphics[width=0.75\linewidth]{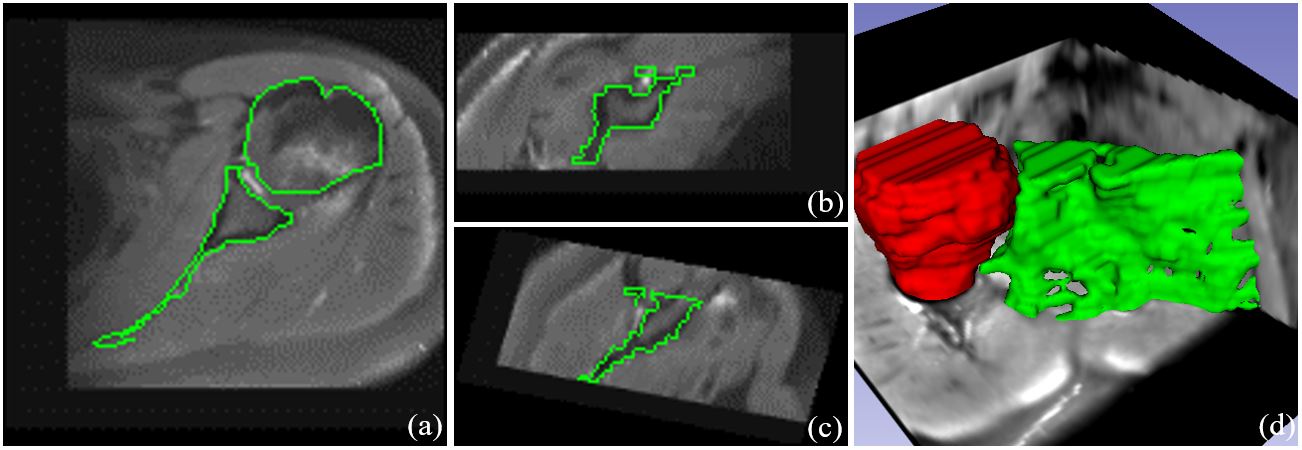} \\
    \includegraphics[width=0.75\linewidth]{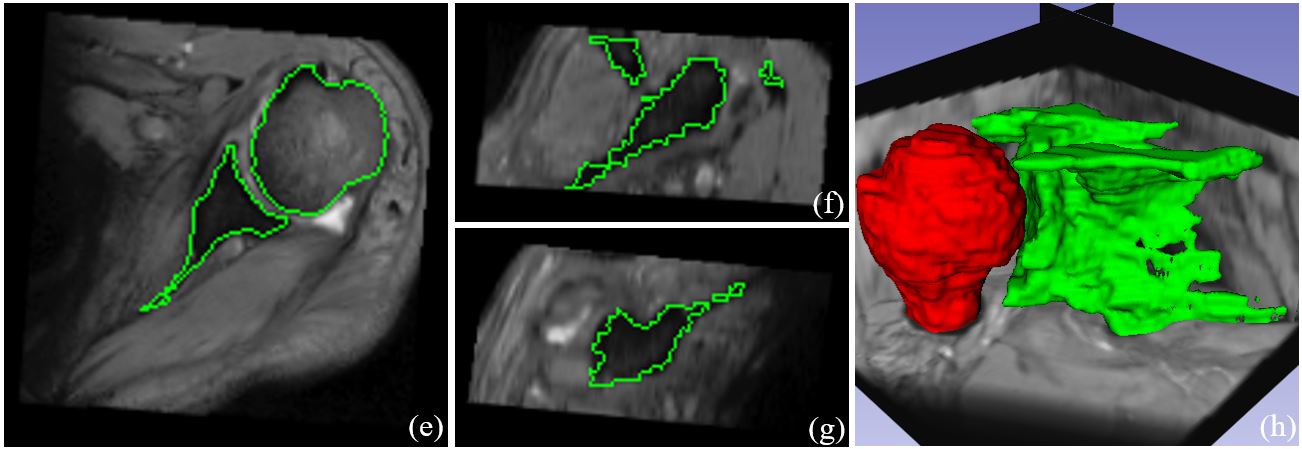}
\end{tabular}
\caption[example]{\label{fig:intro1} Demonstration of the images and GTs from two subjects in the experimental dataset. (a) to (c) show the axial, sagittal and coronal slices from the 3D MR shoulder data of subject one. The green lines show the GT contours. (d) demonstrates the 3D humerus (red) and scapula (green) labels. (e) to (h) show another subject.}
\end{center}
\end{figure}

Some previous automatic and semi-automatic approaches pay more attention to the fine scanned CT shoulder bones images \cite{sharma2013adaptive}. The CT scanning could offer high-contrast bone imaging, but its moderate or high radiation may limit its wide usage for high-resolution and large-region 3D shoulder imaging. On the other hand, due to the heavy demands of MR scanning for shoulder preoperative, obtaining high-resolution (i.e., low slice thickness) MR images are quite impossible, since taking the high-resolution MR scanning is time-consuming in clinical. In addition, the parameters for scanning may differ from case to case, which will increase the variance of images. Fig.~\ref{fig:intro1} exhibits two 3D MR shoulder data from the experimental dataset (with a total of 50 data) of this paper. The two data have complex and diverse bone structure, and their original MR images also have different intensity and contrast. Moreover, all the experimental MR images have been coarsely scanned in the axial direction for fast bone region location, and the images have resolutions between $0.3125\times0.3125\times1.0$ and $0.9615\times0.9615\times4.5$, and sizes between $208\times208\times22$ and $512\times512\times40$. The high divergence of imaging parameters (e.g., resolution, size) and patient conditions (e.g., bone shape) causes the humerus and scapula segmentation problem difficult. Currently, the atlas-based automated extraction approaches and statistical shape models (SSMs) of bony structures build the basic frameworks for scapula and humerus segmentation. An atlas is a pair of an image scan and a corresponding manual labeling mask. The atlas-based segmentation is estimated using image registration, and hence a target image is propagated to single or multiple atlases to extract and fuse the final segmented masks. Wang et al. proposed a representative multi-atlas approach by using joint label fusion \cite{wang2013multi}. This approach computes the atlas voting weights by minimizing a total expected fusion error and it achieves good accuracy. On the development of SSMs, Mutsvangwa et al. reported on an improved pipeline to construct the automated unbiased global SSMs by employing an iterative median closest point-Gaussian mixture model (IMCP-GMM) method \cite{mutsvangwa2014an}. Although both methods can tolerate more variants of shape and appearance of bones, they still highly depend on the availability and proper selection of training samples for the computation of label fusion weights or the generation of mean-virtual shape. Moreover, under the complicated imaging conditions mentioned above, the manually labeled bone ground truth (GT) masks may suffer from the non-continuous boundary issue in 3D space and cause heavy biases to degenerate the segmentation performance of the two types of methods.

\begin{figure}[!t]
    \begin{center}
    \begin{tabular}{c} 
        \includegraphics[height=0.43\textheight]{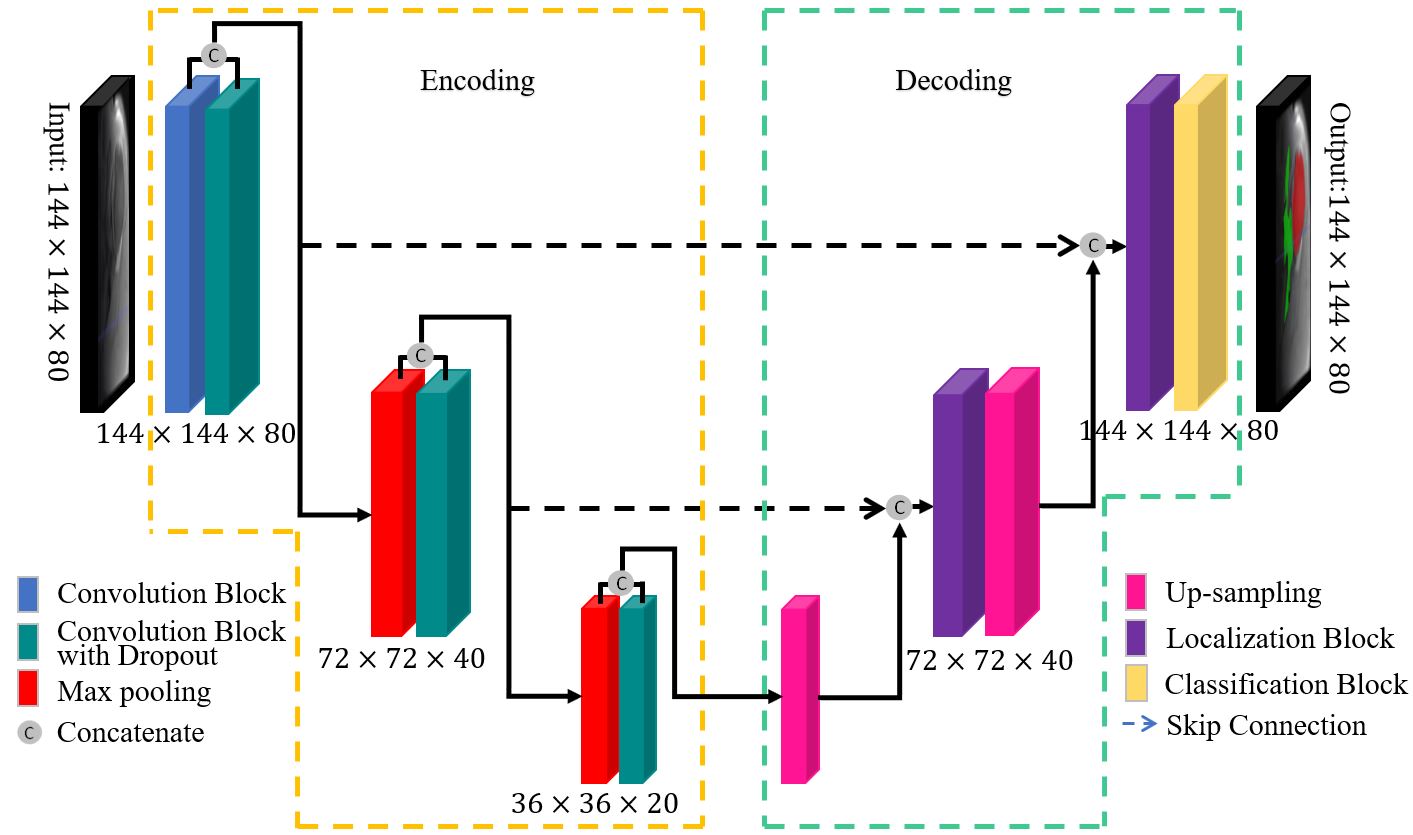}
    \end{tabular}
    \caption[example]{\label{fig:NetworkStructure} Proposed deep end-to-end network. The left half is encoding for the abstraction of multi-level contextual features. The right half is decoding for the bone classification learning. Each convolution block includes one or two convolutional layers with filter size of $3\times3\times3$ and zero-padding of 1. Batch normalization and parametric rectified linear unit are also adopted in all convolutional and deconvolutional layers.}
    \end{center}
\end{figure}

In this paper, we formulate the humerus and scapula segmentation as a deep end-to-end network, and its structure is shown in Fig. \ref{fig:NetworkStructure}. The 3D bone extraction is constructed based on the organ prediction branch task, which is a fully convolutional network for the inference of organ probability map, in the previous work \cite{tan2018deep}. The proposed segmentation architecture exploits the deep convolutional neural networks to infer tissue contextual information on different resolution levels of image and then make a prediction at every pixel. In each resolution stage, the residual connection is also employed to ease the vanishing gradient problem in training when the network goes deeper. Thus, without manually selecting proper reference data or setting handcrafted models, the proposed network has the ability to learn a hierarchical representation of the shape-varied bone data under the coarse imaging conditions. In order to further improve the trained model when using the imperfect GT labels, we introduce a self-reinforced learning strategy that employs the current trained model to generate more and higher quality data to support the next round of model training. Thus, the proposed method achieves an effective bone-marking towards the two important bones (i.e., humerus and scapula) in the coarsely scanned 3D MR knee data, and the results could be used to derive an initial shoulder preoperative diagnosis.

\section{METHOD}
\subsection{Deep end-to-end network for humerus and scapula segmentation}
\label{sec:DE2ENN}

In this section, we present the deep end-to-end network for the bone tissue predication and segmentation. As shown in Fig.~\ref{fig:NetworkStructure}, the main structure of each task is designed as a symmetric encoding-decoding way. The training on the large-size 3D MR shoulder data requires very high GPU memory consumption. Hence, we shorten the depth of both encoding and decoding paths and reduce the number of pooling and deconvolution operations to 2, respectively. In the encoding direction, for each different resolution level, a convolution block is utilized for feature abstraction and followed with a residual connection. The successive multi-resolution encoder obtains multi-size contextual information which is helpful to receive the integral bone structure as well as the background knowledge surrounding the target tissue. In the decoding part, after the skip connection that concatenates the up-sampled low-level feature maps with the skipped equivalent-resolution maps from the encoding half, we utilize a localization block \cite{isensee2017brain} to fuse the concatenated features. The multi-class cross entropy is applied and the loss function is: ${{\cal L}} = - \sum\nolimits_{v \in V} {\sum\nolimits_{j = 1}^K {{t_{v,j}}\log \left( {{p_{v,j}}} \right)} }$. Here $K=3$ is the number of classes, representing the class of humerus, scapula, and background. For the $j$-th class, $p_{v,j}$ and $t_{v,j}$ are the predicted probability and the GT at voxel $v \in$ $V$, where $V$ is the volume data space.

\begin{figure} [!t]
   \begin{center}
   \begin{tabular}{c} 
   \includegraphics[width=0.90\linewidth]{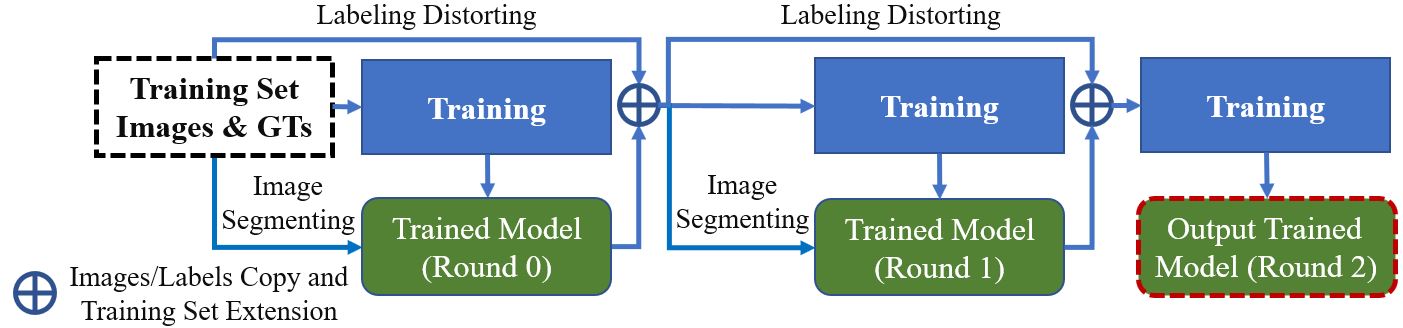}
   \end{tabular}
   \end{center}
   \caption[example] 
   { \label{fig:Reinforced Learning} Flowchart of the self-reinforced learning. After the first training round (i.e., round 0), the trained model and data augmentation techniques (e.g., distortion) are used for generating higher quality data to extend (i.e., $\bigoplus$) the training set to support the subsequent training rounds. The red dashed box represents the improved output model.}
\end{figure}

\subsection{Self-reinforced learning}
\label{sec:Reinforced Learning}
The network in section \ref{sec:DE2ENN} could obtain good initialized bone segmented masks based on the inaccurate GT labels and scarce training data. Yet this network has the potential to raise the segmentation performance. Hence, a self-reinforced learning strategy is proposed and its flowchart is demonstrated as Fig.~\ref{fig:Reinforced Learning}. After the initial training, the trained model and augmentation techniques (e.g., distortion) are utilized to generate higher quality labels and extend the training set, which could be used for the subsequent training rounds. The labels produced by the trained model have more continuous boundaries and are able to represent more precise target regions. Based on our experience, two extra training rounds could make the model converged and have the highest performance.

\section{Experiments}
\subsection{Experiment settings}
\label{sec:Experiment Settings}
In this study, we validate the proposed method on a dataset with 50 3D MR shoulder images. Because of the high divergence of imaging parameters and patient conditions, all the images are resampled and cropped, and have the same voxel spacing ($1mm$, $1mm$, $1mm$) and size ($144\times144\times80$). In addition, the N4 bias field correction is applied to all the data and the pixel intensity is normalized in [0, 1]. In order to validate the effectiveness of the self-reinforced learning, we carry out a 5-folds cross-validation on the experimental dataset. The dataset is randomly split into 5 mutually exclusive sets, and each group chooses four of the five sets for training and use the remaining set for testing. Then we compare the proposed method to the widely used atlas segmentation with joint label fusion (MALF) \cite{wang2013multi}. The proposed network (using the model trained above) and the MALF (using 15 selected atlases as the reference) are implemented using the Tensorflow and Matlab libraries, and the two methods are tested on the same testing dataset. For validation, dice similarity coefficient (DSC), Hausdorff distance (HD) and average surface distance (ASD) between the GT labels and segmented results are reported. In the network training, the batch size is set to 1 and the Adam solver (the learning rate is initialized as $0.001$ and multiplied by a factor of $0.95$ every 10 epochs) is employed.

\subsection{Experiment results}
\label{sec:Experiment Result}
\begin{figure} [!t]
\begin{center}
\setlength{\tabcolsep}{2pt}
   \begin{tabular}{c c} 
        \includegraphics[height=0.20\textheight]{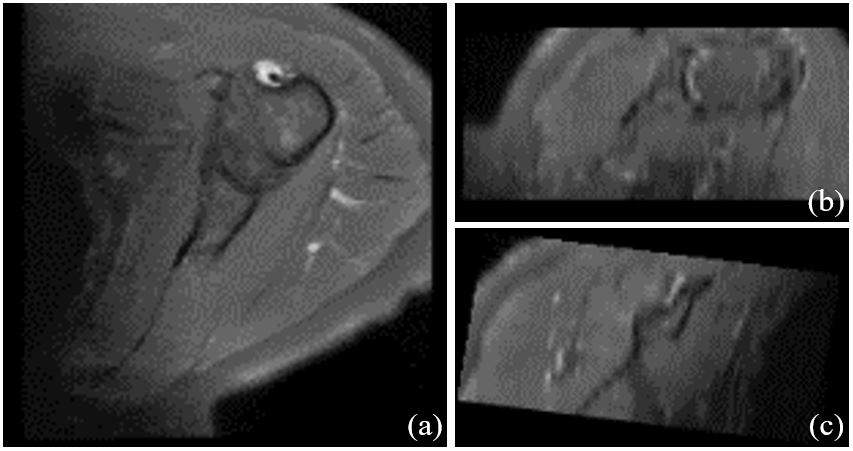} & \includegraphics[height=0.20\textheight]{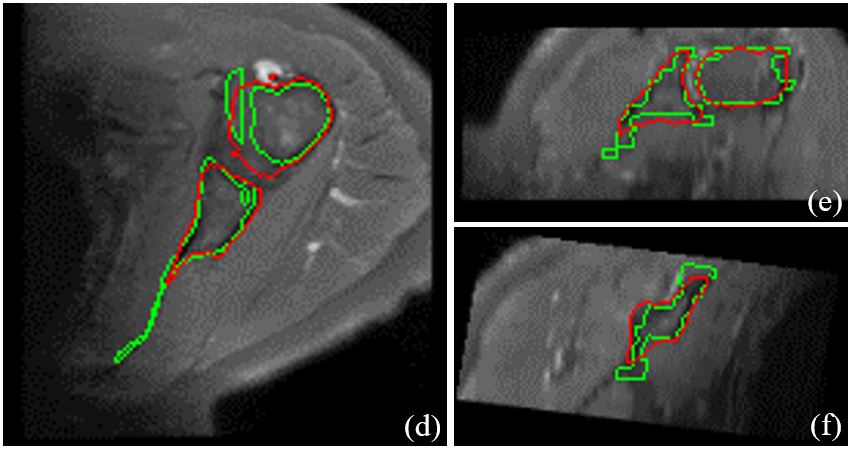} \\
        \includegraphics[height=0.20\textheight]{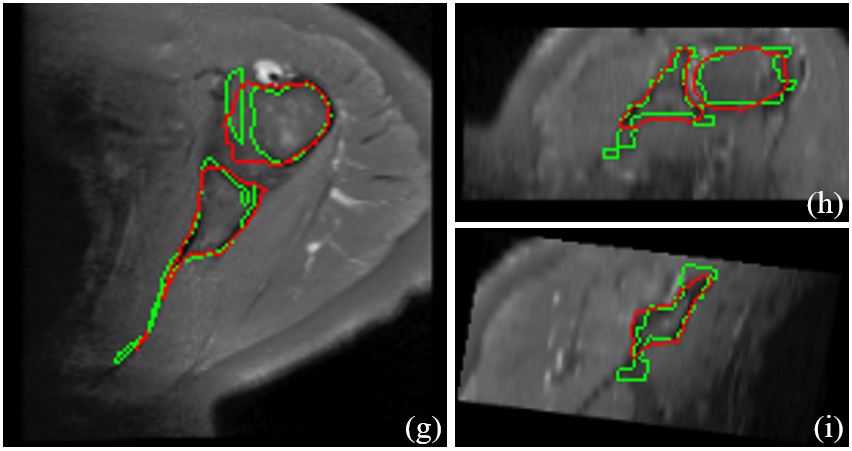}& \includegraphics[height=0.20\textheight]{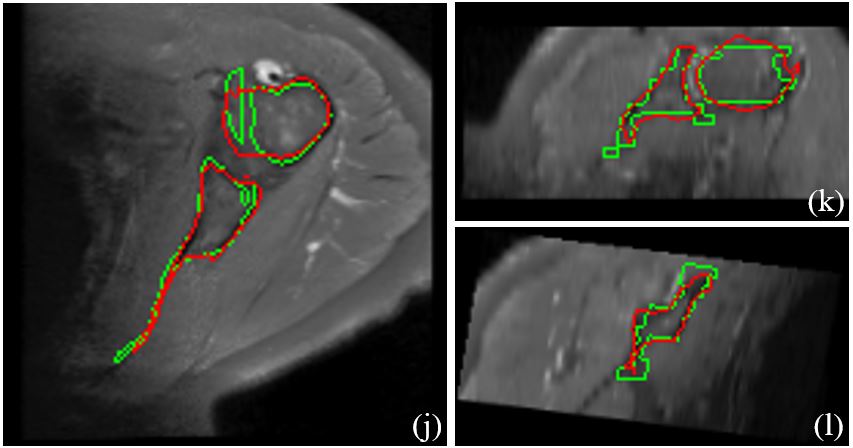}\\
	\end{tabular}
	\end{center}
    \caption[example] 
    { \label{fig:Compare_btw_three_nets} 2D visual comparisons of a subject to validate the self-reinforced learning. Green and red lines are for the GT and segmented results. (a) to (c) show the axial, sagittal and coronal slices of the subject. (d) to (f), (g) to (i), and (j) to (l) demonstrate the progressed segmented bone contours from the training R0, R1 and R2, respectively.}
\end{figure}

\begin{figure} [!t]
\begin{center}
\setlength{\tabcolsep}{1pt}
    \begin{tabular}{c c c c} 
        \includegraphics[height=0.19\textheight]{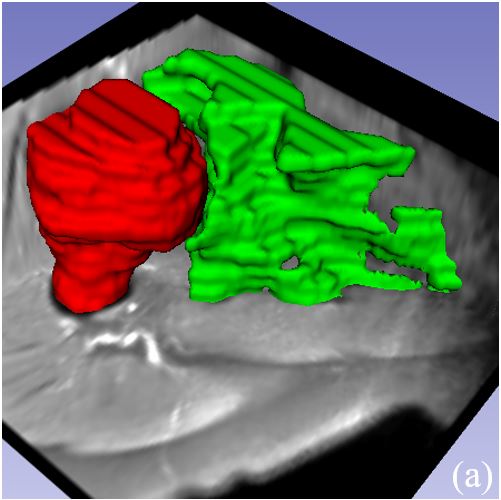} & \includegraphics[height=0.19\textheight]{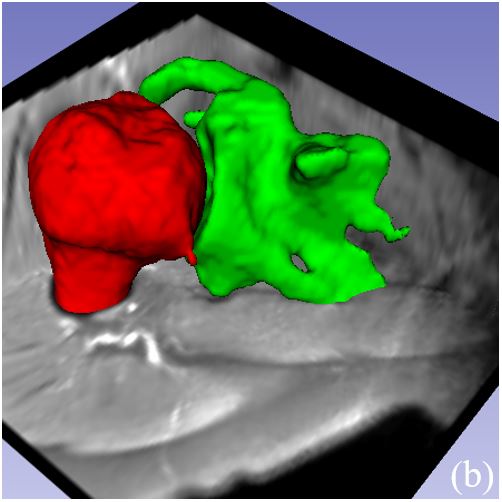} & \includegraphics[height=0.19\textheight]{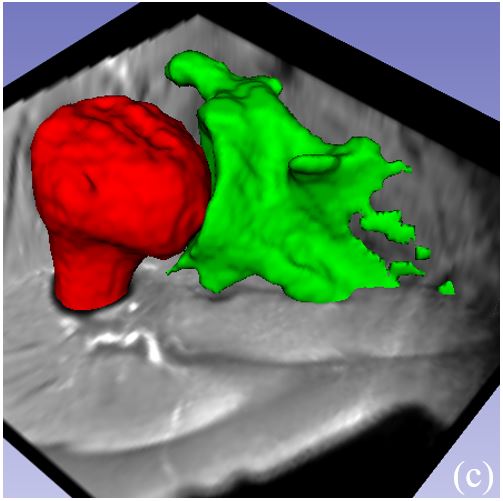} & \includegraphics[height=0.19\textheight]{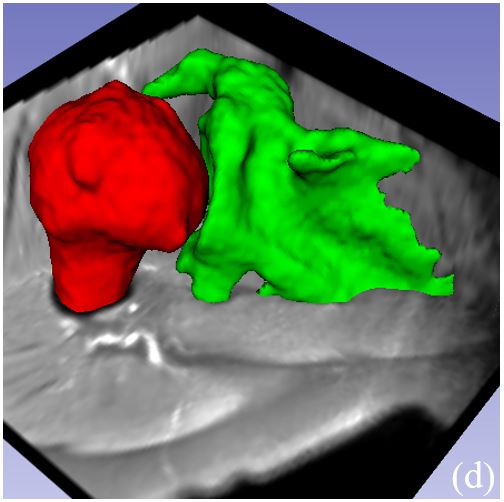}\\
    \end{tabular}
	\end{center}
    \caption[example] 
    { \label{fig:Compare_btw_three_nets_3D} 3D visual comparisons of the same subject in Fig.~\ref{fig:Compare_btw_three_nets}. The red and green labels represent humerus and scapula masks. (a) shows the GT labels, (b) to (d) are the results segmented by the R0, R1, and R2 training models, respectively.}
\end{figure}

Fig.~\ref{fig:Compare_btw_three_nets} shows the 2D visual comparisons of a subject to validate the self-reinforced learning. In Fig.~\ref{fig:Compare_btw_three_nets} (d) to (f), the initialized model from the training round 0 (R0) could locate the main bone areas and describe their basic structures, yet miss some scapula areas. After deploying the self-reinforced learning to optimize the training steps, from the demonstrations in Fig.~\ref{fig:Compare_btw_three_nets} (g) to (i) and (j) to (l), the trained model can progressively segment all the scapula bone and reconstruct more complete humerus and scapula shape. The comprehensive 3D segmentation views of the same subject are also shown in Fig.~\ref{fig:Compare_btw_three_nets_3D}. In Fig.~\ref{fig:Compare_btw_three_nets_3D} (a), the GT labels’ surface and boundary are non-continuous and non-smooth, and these situations even cause some errors of inconsistency between the slices of GT. After the self-reinforced processing, from Fig.~\ref{fig:Compare_btw_three_nets_3D} (c) to (d), the recursively trained models can refine the humerus and scapula masks, as long as smoothing their surface and keeping the details of the bone shapes. The trained model on R2 obtains the best quantitative performance on this subject. The values for humerus are: DSC (0.918), HD (5.099), ASD (0.764), and for scapula are: DSC (0.734), HD (12.329) and ASD (0.784). Besides the comparisons on the single subject, the overall quantitative measurements by using the 5-fold cross-validation are shown in Table~\ref{tab:Group_Comp}. The R2 in the self-reinforced learning almost produces the best mean results among other two rounds in general. Moreover, the shape of humerus has much higher consistency than that of scapula, and thus the prediction of our proposed method on humerus has a higher mean DSC score.
\begin{table}[!t]
    \centering
    \caption[example]{Quantitative comparisons for the self-reinforced learning using 5-fold cross-validation}
    \begin{tabular}{|l|c|c|c|c|c|c|c|c|c|}
        \hline
        \multirow{2}{*}{\textbf{\textit{G1}}} & \multicolumn{3}{c|}{\textbf{Humerus}} & \multicolumn{3}{c|}{\textbf{Scapula}} & \multicolumn{3}{c|}{\textbf{Both}}\\
        \cline{2-10}
        & \textit{DSC} & \textit{HD} & \textit{ASD} & \textit{DSC} & \textit{HD} & \textit{ASD} & \textit{DSC} & \textit{HD} & \textit{ASD} \\
        \hline
        R0 & 0.880 & \textbf{8.030} & 1.285 & 0.685 & 26.685 & 2.138 & 0.809 & 25.468 & 1.727\\
        \hline
        R1 & 0.882 & 8.195 & 1.179 & 0.660 & 25.956 & 1.816 & 0.806 & 25.100 & 1.515\\
        \hline
        R2 & \textbf{0.890} & 8.443 & \textbf{1.119} & \textbf{0.722} & \textbf{21.404} & \textbf{1.581} & \textbf{0.829} & \textbf{18.884} & \textbf{1.363}\\
        \hline
        \hline
        \multirow{2}{*}{\textbf{\textit{G2}}} & \multicolumn{3}{c|}{\textbf{Humerus}} & \multicolumn{3}{c|}{\textbf{Scapula}} & \multicolumn{3}{c|}{\textbf{Both}}\\
        \cline{2-10}
        & \textit{DSC} & \textit{HD} & \textit{ASD} & \textit{DSC} & \textit{HD} & \textit{ASD} & \textit{DSC} & \textit{HD} & \textit{ASD} \\
        \hline
        R0 & 0.866 & 8.579 & 1.327 & 0.606 & 31.184 & 1.588 & 0.784 & 29.642 & 1.506\\
        \hline
        R1 & \textbf{0.876} & \textbf{6.723} & \textbf{1.113} & 0.623 & 29.452 & \textbf{1.194} & 0.800 & 28.555 & \textbf{1.210}\\
        \hline
        R2 & 0.870 & 8.650 & 1.126 & \textbf{0.673} & \textbf{27.024} & 1.500 & \textbf{0.803} & \textbf{24.771} & 1.321\\
        \hline
         \hline
        \multirow{2}{*}{\textbf{\textit{G3}}} & \multicolumn{3}{c|}{\textbf{Humerus}} & \multicolumn{3}{c|}{\textbf{Scapula}} & \multicolumn{3}{c|}{\textbf{Both}}\\
        \cline{2-10}
        & \textit{DSC} & \textit{HD} & \textit{ASD} & \textit{DSC} & \textit{HD} & \textit{ASD} & \textit{DSC} & \textit{HD} & \textit{ASD} \\
        \hline
        R0 & 0.781 & 12.873 & 2.156 & 0.389 & 40.141 & 3.322 & 0.659 & 32.797 & 2.198\\
        \hline
        R1 & 0.833 & 12.281 & 1.659 & 0.546 & 26.993 & \textbf{2.262} & 0.752 & 24.020 & 1.785\\
        \hline
        R2 & \textbf{0.879} & \textbf{10.403} & \textbf{1.203} & \textbf{0.605} & \textbf{19.196} & 2.960 & \textbf{0.797} & \textbf{18.744} & \textbf{1.603}\\
        \hline
        \hline
        \multirow{2}{*}{\textbf{\textit{G4}}} & \multicolumn{3}{c|}{\textbf{Humerus}} & \multicolumn{3}{c|}{\textbf{Scapula}} & \multicolumn{3}{c|}{\textbf{Both}}\\
        \cline{2-10}
        & \textit{DSC} & \textit{HD} & \textit{ASD} & \textit{DSC} & \textit{HD} & \textit{ASD} & \textit{DSC} & \textit{HD} & \textit{ASD} \\
        \hline
        R0 & 0.787 & 13.845 & 2.387 & 0.490 & 34.386 & 1.388 & 0.691 & 32.436 & 1.996\\
        \hline
        R1 & 0.806 & 15.824 & 2.573 & \textbf{0.583} & 34.752 & 1.369 & 0.728 & 33.831 & 2.001\\
        \hline
        R2 & \textbf{0.862} & \textbf{10.498} & \textbf{1.476} & 0.563 & \textbf{35.488} & \textbf{1.079} & \textbf{0.764} & \textbf{30.234} & \textbf{1.330}\\
        \hline
        \hline
        \multirow{2}{*}{\textbf{\textit{G5}}} & \multicolumn{3}{c|}{\textbf{Humerus}} & \multicolumn{3}{c|}{\textbf{Scapula}} & \multicolumn{3}{c|}{\textbf{Both}}\\
        \cline{2-10}
        & \textit{DSC} & \textit{HD} & \textit{ASD} & \textit{DSC} & \textit{HD} & \textit{ASD} & \textit{DSC} & \textit{HD} & \textit{ASD} \\
        \hline
        R0 & 0.867 & 12.346 & 1.928 & 0.674 & 25.341 & 2.357 & 0.802 & 25.368 & 2.002\\
        \hline
        R1 & 0.870 & \textbf{8.611} & 1.256 & 0.645 & 26.483 & \textbf{1.466} & 0.799 & 25.957 & 1.367\\
        \hline
        R2 & \textbf{0.879} & 10.536 & \textbf{1.187} & \textbf{0.723} & \textbf{22.234} & 1.470 & \textbf{0.824} & \textbf{22.031} & \textbf{1.334}\\
        \hline
    \end{tabular}
    \label{tab:Group_Comp}
\end{table}

\begin{figure} [!t]
\begin{center}
\setlength{\tabcolsep}{2pt}
   \begin{tabular}{c c} 
    \includegraphics[height=0.20\textheight]{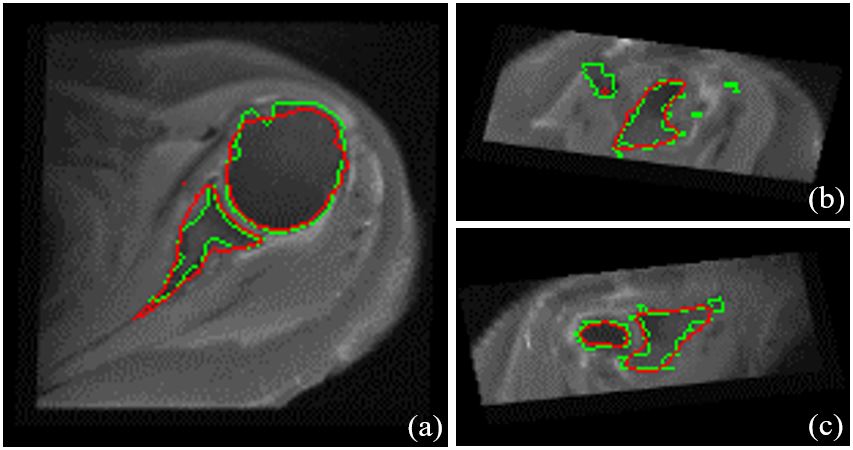} & \includegraphics[height=0.20\textheight]{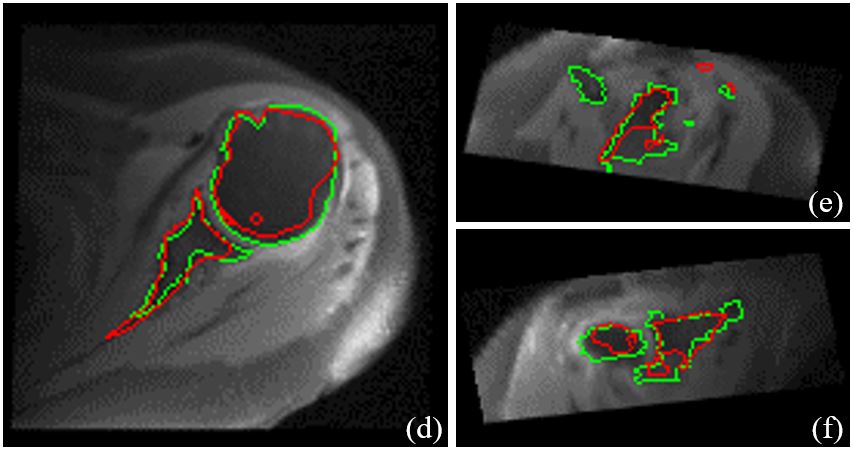}
	\end{tabular}
	\end{center}
    \caption[example] 
    { \label{fig:Compare_btw_two_mthds_lb} 2D visual comparisons of the proposed method and the MALF. Green and red lines are for the GT and segmented results. (a) to (c) show the segmented results of the proposed method, while (d) to (f) are the MALF ones.}
\end{figure}

\begin{figure} [!t]
\begin{center}
    \begin{tabular}{c c c} 
    \includegraphics[width=0.28\textwidth]{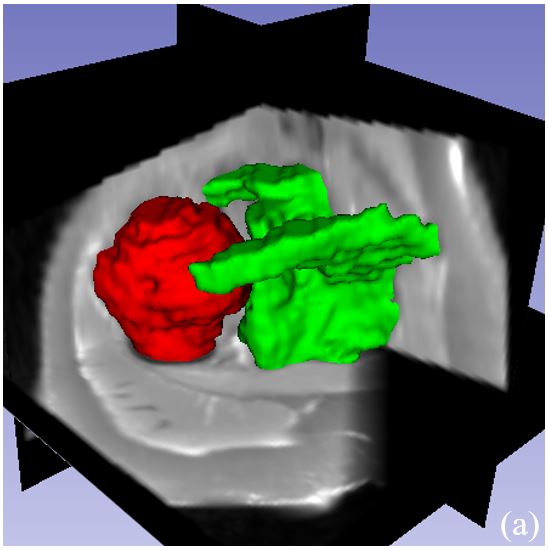} & \includegraphics[width=0.28\textwidth]{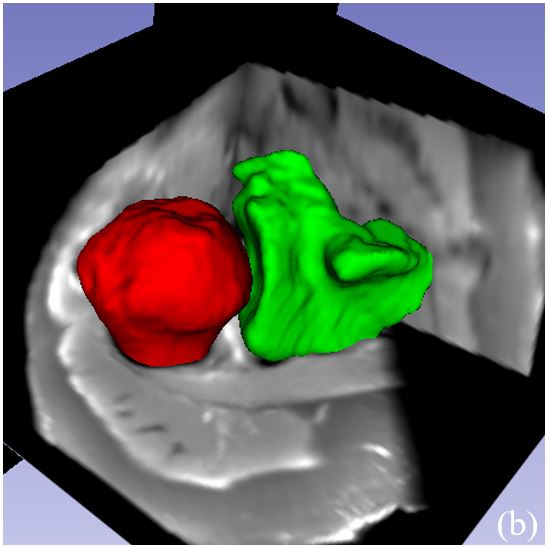} & \includegraphics[width=0.28\textwidth]{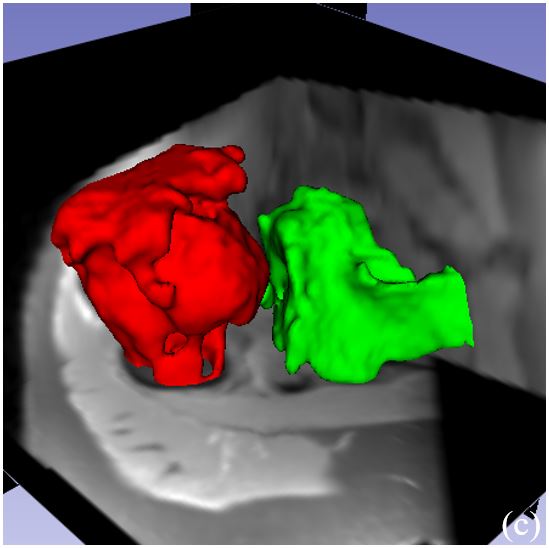}  
	\end{tabular}
	\end{center}
    \caption[example] 
    {\label{fig:Compare_btw_two_mthds_mesh} 3D visual comparisons of the same subject in Fig. \ref{fig:Compare_btw_two_mthds_lb}. The red and green labels represent humerus and scapula masks. (a) shows the GT labels. (b) is the proposed method’s results. (c) is segmented by the MALF.}
\end{figure}

After showing the effectiveness of the proposed network and self-reinforced learning in refining the continuousness and smoothness on the segmented structures, we also visually compare the performance of the proposed method with the widely used MALF approach. By comparing the results in Fig.~\ref{fig:Compare_btw_two_mthds_lb} and~\ref{fig:Compare_btw_two_mthds_mesh}, both methods are able to locate the two bones, and the proposed one can overlap more areas with the GT. The MALF produces some rough “hole” defects on the segmented bone surfaces and also generates several spatially isolated segmentation errors. Moreover, the MALF's 3D masks have obvious leaking issues on the humerus and simultaneously it misses some bone tissue in the scapula part. All of the above problems in the MALF may be caused by the improper reference atlas pairs selection, and the weak appearance information and various anatomical structures in the experimental data may make the similarity estimation inaccurate in the joint fusion step. On the other hand, our method could alleviate the isolated labeling errors and preserve the spiny shape of the scapula. In Table~\ref{tab:Mthd_Comp}, we report the mean evaluation metrics, and the proposed one outperforms the MALF. Thus, the proposed method has higher robustness to segment the small dataset with low-contrast and high-shape-variability 3D MR Data.

\begin{table}[!t]
    \centering
    \caption[example]{Quantitative comparisons between the proposed method and MALF}
    \begin{tabular}{|l|c|c|c|c|c|c|c|c|c|}
        \hline
        \multirow{2}{*}{} & \multicolumn{3}{c|}{\textbf{Humerus}} & \multicolumn{3}{c|}{\textbf{Scapula}} & \multicolumn{3}{c|}{\textbf{Both}}\\
        \cline{2-10}
        & \textit{DSC} & \textit{HD} & \textit{ASD} & \textit{DSC} & \textit{HD} & \textit{ASD} & \textit{DSC} & \textit{HD} & \textit{ASD} \\
        \hline
        MALF & 0.651 & 16.754 & 4.228 & 0.455 & 34.624 & 2.091 & 0.608 & 31.595 & 3.033 \\
        \hline
        \textbf{Our} & \textbf{0.923} & \textbf{4.827} & \textbf{0.715} & \textbf{0.753} & \textbf{21.425} & \textbf{1.272} & \textbf{0.859} & \textbf{20.413} & \textbf{1.019} \\
        \hline
    \end{tabular}
    \label{tab:Mthd_Comp}
\end{table}

\section{Conclusion}
In the present work, we propose a deep end-to-end network and a self-reinforced learning strategy to segment humerus and scapula using low-contrast and high-shape-variability 3D MR shoulder data. The network has a U-shaped structure with an encoding-decoding architecture to formulate the bone extraction as a semantic segmentation with deep learning. In order to further improve the segmentation accuracy and ensure the continuousness and smoothness in the segmented structures, we introduce the self-reinforced learning mechanism. By starting from the small dataset with inaccurate GT labels, this process utilizes the initialized segmentation model to recursively extend the training set with newly generated higher quality labels and improve the next-round training. In the experiments, the proposed method achieves accurate segmentation evaluated with a 5-folds cross-validation and has superior performance by comparing with the MALF approach.
\bibliography{report} 

\begin{thebibliography}{1}

\bibitem{chuang2008use}
Chuang, T.-Y., Adams, C.~R., and Burkhart, S.~S., ``Use of preoperative
  three-dimensional computed tomography to quantify glenoid bone loss in
  shoulder instability,'' {\em Arthroscopy: The Journal of Arthroscopic \&
  Related Surgery}~{\bf 24}(4),  376--382 (2008).

\bibitem{acid2012preoperative}
Acid, S., Le~Corroller, T., Aswad, R., Pauly, V., and Champsaur, P.,
  ``Preoperative imaging of anterior shoulder instability: diagnostic
  effectiveness of mdct arthrography and comparison with mr arthrography and
  arthroscopy,'' {\em American Journal of Roentgenology}~{\bf 198}(3),
  661--667 (2012).

\bibitem{sharma2013adaptive}
Sharma, G.~B. and Robertson, D.~D., ``Adaptive scapula bone remodeling
  computational simulation: Relevance to regenerative medicine,'' {\em Journal
  of Computational Physics}~{\bf 244},  312--320 (2013).

\bibitem{wang2013multi}
Wang, H., Suh, J.~W., Das, S.~R., Pluta, J.~B., Craige, C., and Yushkevich,
  P.~A., ``Multi-atlas segmentation with joint label fusion,'' {\em IEEE
  Transactions on Pattern Analysis and Machine Intelligence}~{\bf 35},
  611--623 (March 2013).

\bibitem{mutsvangwa2014an}
Mutsvangwa, T., Burdin, V., Schwartz, C., and Roux, C., ``An automated
  statistical shape model developmental pipeline: Application to the human
  scapula and humerus,'' {\em IEEE Transactions on Biomedical Engineering}~{\bf
  62},  1098--1107 (April 2015).

\bibitem{tan2018deep}
Tan, C., Zhao, L., Yan, Z., Li, K., Metaxas, D., and Zhan, Y., ``Deep
  multi-task and task-specific feature learning network for robust shape
  preserved organ segmentation,'' in [{\em Biomedical Imaging (ISBI 2018), 2018
  IEEE 15th International Symposium on}{\nolinebreak\hspace{0.1em}]},
  1221--1224, IEEE (2018).

\bibitem{isensee2017brain}
Isensee, F., Kickingereder, P., Wick, W., Bendszus, M., and Maier-Hein, K.~H.,
  ``Brain tumor segmentation and radiomics survival prediction: Contribution to
  the brats 2017 challenge,'' in [{\em International MICCAI Brainlesion
  Workshop}{\nolinebreak\hspace{0.1em}]},   287--297, Springer (2017).

\end{thebibliography}
\bibliographystyle{spiebib} 

\end{document}